\documentclass[acmsmall]{acmart}
\usepackage{graphicx}
\begin{document}

\title{\textcolor{red}{Tox}\textcolor{teal}{Vis}: Enabling Interpretability of Implicit vs. Explicit Toxicity
Detection Models with Interactive Visualization}



\author{Uma Gunturi*}
\affiliation{%
    \authornote{Both authors contributed equally to this research.}
  \institution{Department of Computer Science, Virginia Tech}
  \city{Blacksburg}
  \country{USA}}
\email{umasushmitha@vt.edu}

\author{Xiaohan Ding*}
\affiliation{%
  \institution{Department of Computer Science, Virginia Tech}
  \city{Blacksburg}
  \country{USA}}
\email{xiaohan@vt.edu}

\author{Eugenia H. Rho}
\affiliation{%
  \institution{Department of Computer Science, Virginia Tech}
  \city{Blacksburg}
  \country{USA}}
\email{eugenia@vt.edu}

 \maketitle
 \textcolor{orange}{\textbf{WARNING:}} This paper contains some content which is offensive in nature.
\section{Introduction}
 
 People’s judgment of hateful online content is inherently subjective and multi-faceted as interpretations around what is and is not hateful can depend on personal values, social identities, and culture \cite{nockleby2000hate}. Hence, as opposed to explicit hate-speech or profanity-laden attacks where the offender’s mal intent is expressed in black and white language, it is more challenging to effectively respond to, or mitigate harms caused by \textit{implicit} hatespeech. This is because implicit toxicity often takes form as humour \cite{fortuna2018survey}, insider expressions,  neologisms \cite{cscw-humour}, and microaggressions, \cite{Espaillat2019AnES},  (e.g. "Black people seem to think everything revolves around them being royalty" (from r/unpopularopinion). Hence, human moderators and content moderation systems have difficulty understanding, recognizing, and responding to implicit hatespeech \cite{wright2021recast}, resulting in moderation failures where false negatives (harmful, yet undetected texts) are left unmoderated \cite{10.1145/3479610}. Furthermore, most hate-classification research, including those that examine implicit hatespeech, often use black-box deep learning (DL) models that lack interpretability, failing to explain why a content was classified as hateful. In this work, we aim to address these challenges by developing a visually interactive and explainable tool called \textit{ToxVis} (Fig.1). We built ToxVis by using RoBERTa [cite], XLNET [cite], and GPT-3 [cite] and fine-tuning two transformer-based models to classify hatespeech into three categories of hatespeech: implicit, explicit, and non-hateful. We then used DL interpretation techniques to make the classification results explainable. Through ToxVis, users can \textbf{(1)} type in a potentially hateful text into the system. The system will then \textbf{(2)} classify the input text into one of the three categories of hatespeech (implicit, explicit, or non-hateful). The user can then click on the classification results  \textbf{(3)} to see which words from the input text contributed most to the classification decision, as well as the model's prediction confidence score. Live demo of ToxVis is available \href{https://tinyurl.com/4zybmwkx}{\color{blue}{here}}.

\section{ToxVis Interactive System}
\textbf{Data collection and pre-processing}. We collected existing hate speech datasets from multiple sources: ETHOS  \cite{mollas2020ethos}, AbuseEval \cite{caselli-etal-2020-feel}, and a microaggressions dataset from Tumblr \cite{breitfeller-etal-2019-finding}. The combined corpora had multiple labeling structures (implicit vs. explicit; hate vs. non-hate). Hence, we limited our sample to those that contained explicit, implicit, and non-hate labels. Three researchers then manually verified a randomized sample into three categories: explicit hate, implicit hate, and non-hate. Through this process, we eliminated previously mislabeled instances. We further eliminated instances with urls, and special characters, leaving us with 14,473 posts. 

\hspace{-4mm}\textbf{Training and interpretation}. We fine-tuned transformer-based models (BERT, RoBERTa, XLNET, GPT-3) on our data (train:dev:test split of 33:33:33) for a multi-class classification task (label 1: \textit{explicit hate}, label 2: \textit{implicit hate}, label 3: \textit{non-hate}). We implemented the models using the HuggingFace Transformers \cite{wolf_huggingfaces_2020} and PyTorch libraries \cite{NEURIPS2019_9015}. We then evaluated the fine-tuned models on a held-out test set using standard evaluation metrics (accuracy, precision, recall, and F1). We further improved model performance through hyperparameter tuning with multiple learning rates ($lr = $ 1e-5, 2e-5 or 3e-5), batch sizes ($n_{bs} = $ 16, 32, or 64), and the number of epochs ($n_{epoch} = $ $3$, $4$, or $5$). 
\begin{figure}[h]
\centering
   \includegraphics[width=0.92\textwidth]{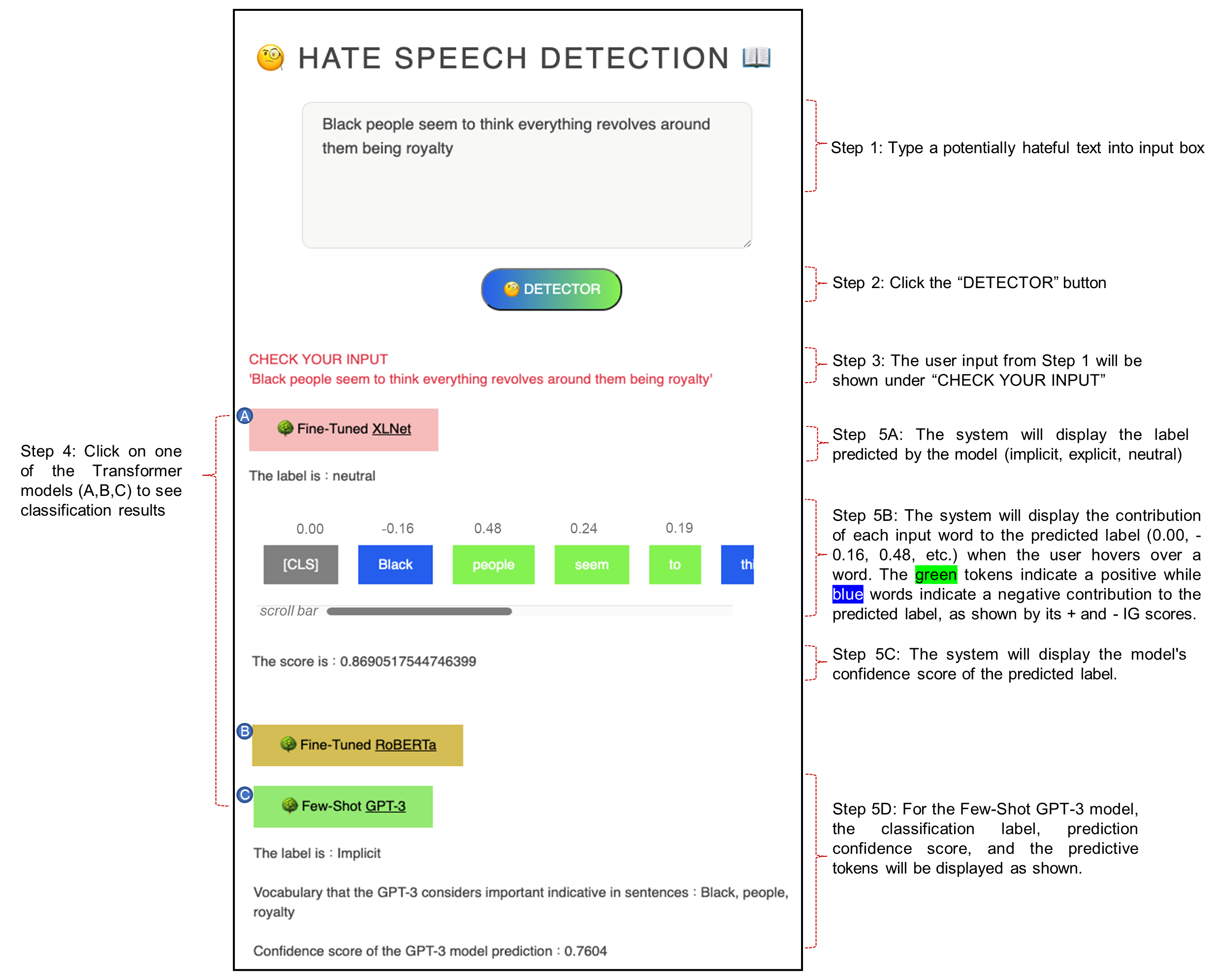}
   \caption{The \textit{ToxVis} user interface. The system leverages three different Transformer-based models A. XLNet B. RoBERTa C. OpenAI's GPT-3. For A and B, the system displays  (1) the model's predicted label (explicit, implicit, non-hateful), (2) the contribution of each input word to the predicted label, and (3) the model's confidence score of the predicted label. For C, the system displays (1) the classification label, (2) the prediction confidence score, and (3) the predictive tokens.}
   \label{fig:teaser}
\end{figure}
\vspace{-2mm}
To make the classification results explainable, we used Integrated Gradients (IG) \cite{sundararajan_axiomatic_2017}, a post-hoc model interpretability technique, to investigate the contribution of individual tokens to the classified label. In other words, we wanted to understand how much a given word from the input text contributed to whether the text was classified as 1. explicitly hateful, 2. implicitly hateful, or 3. non-hateful. For this task, we used IG to calculate the gradient of each input token to analyze the attribution score (ranging from -1 to 1) on the final label. Positive attribution scores indicate a given token's positive contribution towards the predicted label, whereas negative attribution values of a given token do not contribute  to the predicted label. 

\begin{table}[!h]
\centering
\caption{Comparison of F1 score, Precision, Recall, and Accuracy among different language models.}
\label{tab:langmodels}
\scalebox{0.8}{
\begin{tabular}{|c|c|c|c|c|}
\hline $\begin{gathered}\text { Model }
\text { Name }\end{gathered}$ & F1 Score & Precision & Recall & Accuracy \\
\hline BERT & $0.750$ & $0.724$ & $0.778$ & $0.755$ \\
\hline RoBERTa & $0.848$ & $0.829$ & $0.869$ & $0.820$ \\
\hline XLNet & $0.804$ & $0.745$ & $0.873$ & $0.791$ \\
\hline $\begin{gathered}\text { GPT-3 } 
\text { (davinci) }\end{gathered}$ & $0.976$ & $0.981$ & $0.971$ & $0.977$ \\
\hline
\end{tabular}}
\end{table}


\hspace{-4mm} \textbf{Interactive System}. After completing the training and evaluation processes, we deployed our explainable multi-label classification system on a cloud server utilizing Python Flask. The user-interface design goals of ToxVis (Fig.1) was to empower users by allowing them to interactively compare the classificaiton results of three different linguistic DL classifiers. We also visualized the input tokens by different colors according to their contribution to the output prediction: \textcolor{blue}{blue} indicating negative and \textcolor{teal}{green} indicating positive contribution to the predicted labels. We instructed GPT-3 to identify and list words from the input sentence that contributed most (in descending order of importance) in conveying the harmful aspect of the input sentence. 


\bibliographystyle{ACM-Reference-Format}
\bibliography{sample-base}

\end{document}